\documentclass[journal,twocolumn,letterpaper]{IEEEJERM}
\usepackage{cite}
\ifCLASSINFOpdf
\else
\fi

\usepackage{times,amsmath,epsfig}
\usepackage{fancyhdr}
\usepackage{amsmath}
\usepackage{amsfonts}
\usepackage{amssymb}
\usepackage[latin1]{inputenc}
\usepackage{array}
\usepackage{graphicx}
\usepackage{lineno,hyperref}
\usepackage[justification=centering]{caption}
\usepackage{url}
\usepackage{subfigure}
\usepackage{bm}
\usepackage{breqn}
\usepackage{xcolor}
\usepackage{soul}
\usepackage{amssymb}
\usepackage{booktabs}
\usepackage{multirow}

\begin{document}

\title{A Review of Recent Advances of Binary Neural Networks for Edge Computing}

\author{Wenyu~Zhao,~\IEEEmembership{}
        Teli~Ma,~\IEEEmembership{}
        Xuan~Gong,~\IEEEmembership{}
        Baochang~Zhang*,~\IEEEmembership{Member,~IEEE,}
        and~David~Doermann,~\IEEEmembership{Fellow,~IEEE}% <-this % stops a space
}% <-this % stops a space

\markboth{IEEE Journal on Miniaturization for Air and Space Systems}
{Wenyu~Zhao, 
        Teli~Ma, 
        Xuan~Gong, 
        Baochang~Zhang, 
        and~David~Doermann,  \MakeLowercase{\textit{et al.}}: IEEE Journal on Miniaturization for Air and Space Systems}

\twocolumn[
\begin{@twocolumnfalse}
  
\maketitle

\begin{abstract}
Edge computing is promising to become one of the next hottest topics in artificial intelligence because it benefits various evolving domains such as real-time unmanned aerial systems, industrial applications, and the demand for privacy protection. This paper reviews recent advances on binary neural network (BNN) and 1-bit CNN technologies that are well suitable for front-end, edge-based computing. We introduce and summarize existing work and classify them based on gradient approximation, quantization, architecture, loss functions, optimization method, and binary neural architecture search. We also introduce applications in the areas of computer vision and speech recognition and discuss future applications for edge computing.
\end{abstract}


\end{@twocolumnfalse}]

% Put footnotes here
{
  \renewcommand{\thefootnote}{}%
  \footnotetext[1]{Wenyu Zhao, Teli Ma and Baochang Zhang are with Beihang University.}
  \footnotetext[2]{Xuan Gong and David Doermann are with University at Buffalo.}
  \footnotetext[3]{* Corresponding author, bczhang@139.com}
}
 
% \fi
\IEEEpeerreviewmaketitle

\section{Introduction}
\IEEEPARstart{W}{ith} the rapid development of information technology, cloud computing with centralized data processing cannot meet the needs of applications that require the processing of massive amounts of data, nor can they be effectively used when privacy requires the data to remain at the source. Thus, edge computing has become an alternative to handle the data from front-end or embedded devices. Edge computing is a generic term that refers to the use of network computing, and storage capabilities to directly provide services for end-users.  Intelligent edge devices benefit many requirements within real-time unmanned aerial systems, industrial systems, and privacy-preserving applications. Yet convolutional neural networks (CNNs), which have become increasingly important because of their superior performance, suffer from a large memory footprint and high computational cost that makes them difficult to deploy on such devices. For example, in unmanned systems, UAVs serve as a computing terminal with limited memory and computing resources, so it is difficult to complete real-time data processing based on CNNs. To improve the storage and computation efficiency, binary neural networks (BNNs) have shown promise for practical applications.  Binary neural networks are neural networks where the weights are binarized. 1-bit CNNs are an extremely compressed version of BNNs that binarize both the weights and activations to further decrease the model size and the computational cost. These highly compressed models make them suitable for front-end computing. In addition to these two, there are also other quantizing neural networks, such as pruning and sparse neural network widely used in the field of edge computing.

We note that Qin et al. \cite{qin2020binary} review binarization algorithms in terms of  minimizing the quantization error, improving the loss function, or reducing gradient error. Different from it, we review BNN methods in a more elaborated manner, including binary neural architecture search which is not mentioned before and more applications.

In this survey, we will review the main advancements of binary neural networks and 1-bit CNNs.  Although binarizing operations can make neural networks more efficient, they almost always cause a significant performance drop. In the last five years, many methods have been introduced to improve the performance of binary neural networks. To better review these methods, we six aspects including gradient approximation, quantization, structural design, loss design, optimization, and binary neural architecture search. Finally, we will also review object detection, object tracking, and audio analysis applications of BNNs.

% \hfill Z. Peng
 
% \hfill Jun 30, 2017

\section{Principal Methods}
In this section, we will review binary neural networks and 1-bit neural networks and highlight similarities and differences between them.

\begin{table*}[t]
			\caption{Results Reported in BinaryConnect  \cite{courbariaux2015binaryconnect} and BinaryNet \cite{courbariaux2016binarized}}
			\label{table1}
			\centering
			\begin{tabular}{c c c}
				\toprule
				Method & MNIST & CIFAR-10\\
				\hline
				\hline
				BinaryConnect (only binary weights) & 1.29$\pm$0.08\% & 9.90\% \\
				BinaryNet (binary both weights and activations) & 1.40\% & 10.15\% \\
				\bottomrule
			\end{tabular}
		\end{table*}

\subsection{Early Binary Neural Networks}
\label{sec2.1}
BinaryConnect \cite{courbariaux2015binaryconnect} was the first work presented that tried to constrain the weights to either +1 or -1 during propagations, but it does not binarize the inputs. The binary operations are simple and readily comprehensible. One way to binarize CNNs is by using a sign function:
\begin{equation}
	\label{eq1}
\omega_b =\left\{
           \begin{array}{lcl}
     +1,\quad   &if \quad \omega\geq0 \\
    -1,\quad  &otherwise 
           \end{array},
        \right.
	\end{equation}
where $\omega_b$ is the binarized weight and $\omega$ the real-valued weight. A second way is to binarize stochastically:

\begin{equation}
	\label{eq2}
\omega_b =\left\{
           \begin{array}{lcl}
     +1,\quad   &with \quad probability \quad p=\sigma(\omega) \\
    -1,\quad  &with \quad probability \quad 1-p 
           \end{array},
        \right.
	\end{equation}
where $\sigma$ is the "hard sigmoid" function.
The training process for these networks is slightly different from full-precision neural networks. The forward propagation utilizes the binarized weights instead of the full-precision weights, but the backward propagation is the same as conventional methods. The gradient $\frac{\partial C}{\partial \omega_b}$ needs to be calculated ($C$ is the cost function) and then combined with the learning rate to update the full-precision weights directly.

BinaryConnect only binarizes the weights, while BinaryNet \cite{courbariaux2016binarized} quantizes both the weights and activations. BinaryNet also introduces two ways to constrain weights and activations to be either +1 or -1, like BinaryConnect. BinaryNet also makes several changes to adapt to binary activations. The first is to shift based Batch Normalization (SBN), which avoids additional multiplications. The second is to shift based AdaMax instead of ADAM learning rule, which also decreases the number of multiplications. The one-third change is to the operation to the input of the first layer. BinaryNet handles continuous-valued inputs of the first layer as fixed point numbers, with $\textit{m}$ bits of precision. Training neural networks with extremely low bit weights and activations were proposed as QNN \cite{hubara2017quantized}. As we are primarily reviewing work on binary networks, the details of QNN are omitted here.  The error rates of these networks on representative datasets are shown in Table \ref{table1}. However, on larger datasets, these two networks perform unsatisfactorily, since weights constrained to +1 and -1 cannot be learned effectively. New methods for training binary neural networks and 1-bit networks need to be raised.

Song Wang et al. \cite{wang2016study} proposed Binarized Deep Neural Networks (BDNNs) for image classification tasks, where all the values and operations in the network are binarized. While BinaryNet deals with convolutional neural networks, BDNNs target at basic artificial neural networks which consists of full-connection layers. Bitwise neural networks \cite{KimS16} also present a completely bitwise network where all participating variables are bipolar binaries. 

\subsection{Gradient Approximation}
As described in Section \ref{sec2.1}, while updating the parameters in BNNs and 1-bit networks, the full-precision weights are updated with the gradient $\frac{\partial C}{\partial \omega_b}$. But in the forward propagation, there is a sign function between full-precision weights and binarized weights. In other words, the gradient of sign function should be considered when updating full-precision weights. Note that the derivative of sign function keeps zero and only becomes infinity at zero points, and a derivable function is widely utilized to approximate the sign function.  

The first one to solve this problem in a 1-bit network is BinaryNet \cite{courbariaux2016binarized}. Assuming that an estimator of $g_q$ of the gradient $\frac{\partial C}{\partial q}$, where $q$ is $Sign(r)$, has been obtained. Then, the straight-through estimator of $\frac{\partial C}{\partial r}$ is simply

\begin{equation}
    g_r = g_q1_{\left|r\right|\leq1},
\end{equation}
where $1_{\left|r\right|\leq1}$ equals to $1$ when $\left|r\right|\leq1$. And it equals to $0$ in other cases. It can also be seen as propagating the gradient through hard $\textit{tanh}$, which is a piece-wise linear activation function.

Bi-real Net \cite{liu2018bi} approximates the derivative of the sign function for activations. Different from using $\textit{Htanh}$ \cite{courbariaux2016binarized} to approximate sign function, Bi-real Net uses a piecewise polynomial function for a better approximation. 

Bi-real Net also proposes a magnitude-aware gradient for weights. While training the binary neural networks, the gradient $\frac{\partial C}{\partial W}$ is only related to the sign of weights, and it is independent of its magnitude. So Bi-real Net replaces the sign function by a magnitude-aware function.

Xu et al. \cite{xu2019accurate} use a higher-order approximation for weight binarization. They propose a long-tailed approximation for activation binarization as a trade-off between tight approximation and smooth backpropagation.

DSQ \cite{gong2019differentiable} also introduces a function to approximate the standard binary and uniform quantization process, which is called differentiable soft quantization. DSQ employs a series of hyperbolic tangent functions to gradually approach the staircase function for low-bit quantization (sign function in 1-bit CNN). The binary DSQ function is as follows:

\begin{equation}
	\label{eq5}
Q_s(x) =\left\{
           \begin{array}{lcl}
     -1,\quad   & x<-1 \\
    1,\quad  & x>1 \\
    stanh(kx), & otherwise
           \end{array},
        \right.
	\end{equation}
with
\begin{equation}
    k=\frac{1}{2}log(\frac{2}{\alpha}-1), s=\frac{1}{1-\alpha}.
\end{equation}

Especially, when $\alpha$ is small, DSQ can well approximate the performance of the uniform quantization. This means that an appropriate $\alpha$ will enable DSQ to help train a quantized model with higher accuracy. Note that DSQ is differentiable, thus the derivative of this function can be used while updating parameters directly.

According to these methods above, we can summarize that they all introduce a new function which is differentiable to approximate the sign function in BinaryConnect so that the gradient to full-precision weights or activations can be obtained more accurately. Therefore, in the training process, the binary neural network or 1-bit network converge easier, the performance of the network improves as well.
\subsection{Quantization}
BinaryConnect and BinaryNet, use simple quantization methods. After the full-precision weights are updated, the new binary weights are generated by taking the sign of real-value weights. But when the binary weights are decided only by the sign of full-precision weights, this may cause significant errors in quantization. Before introducing new methods to improve the quantization process, we highlight the notations used in XNOR-Net \cite{ rastegari2016xnor} that will be used in our discussions. For each layer in a convolutional neural network, $\textit{I}$ is the input, $\textit{W}$ is the weight filter, $\textit{B}$ is the binarized weight (+1 and -1) and $\textit{H}$ is the binarized input.

Rastegari et al. \cite{rastegari2016xnor} propose Binary-Weight-Networks (BWN) and XNOR-Networks. BWN approximates the weights with binary values a variation of a binary neural network. XNOR-Networks binarize both the weights and activations bit is considered a 1-bit network. Both networks use the idea of a scaling factor. In BWN, the real-value weight filter $\textit{W}$ is estimated using a binary filter $\textit{B}$ and a scaling factor $\alpha$. The convolutional operation is then approximated by:

\begin{equation}
  I*W \approx (I \oplus B)\alpha,  
\end{equation}
where $\oplus$ indicates a convolution without any multiplication. By introducing the scaling factor, the binary weight filters reduce memory usage by a factor of $32\times$ compared to single precision filters. To ensure $\textit{W}$ is approximately equal to $\alpha \textit{B}$, BWN defines an optimization problem, and the optimal solution is:

\begin{equation}
    B^* = sign(W),
\end{equation}

\begin{equation}
    \alpha^* = \frac{W^T sign(W)}{n}=\frac{\sum \left|W_i\right|}{n}=\frac{1}{n} \|W_r\|_{l_1}.
\end{equation}

The optimal estimation of a binary weight filter can thus be achieved by taking the sign of weight values. The optimal scaling factor is the absolute average of absolute weight values. The scaling factor is also used in calculating the gradient in backpropagation. The core idea of XNOR-Net is the same as BWN, but another scaling factor $\beta$ is used while binarizing input $I$ into $H$. As experiments show, this approach outperforms BinaryConnect and BNN by a large margin on ImageNet. Unlike the XNOR-Net which sets the mean of weights as the scaling factor, Xu et al. \cite{xu2019accurate} define a trainable scaring factor for both weights and activations. LQ-Nets \cite{zhang2018lq} quantize both the weights and the activations with arbitrary bit-widths, including 1-bit width. The learnability of the quantizers makes them compatible with bitwise operations to keep the fast inference merit of properly-quantized neural networks.

Based on XNOR-Net \cite{rastegari2016xnor}, HORQ \cite{li2017performance} provides a high-order binarization scheme, which achieves a more accurate approximation while still having the advantage of the binary operation. HORQ calculates the residual error and then performs a new round of thresholding operations to further approximate the residual. This binary approximation of the residual can be considered a higher-order binary input. Following XNOR, HORQ defines the first-order residual tensor $R_1(x)$ by computing the difference between the real input and first-order binary quantization:

\begin{equation}
    R_1(x) = X-\beta_1 H_1 \approx \beta_2 H_2,
\end{equation}
where $R_1(x)$ is a real value tensor.  By this analogy, $R_2(x)$ can be seen as the second-order residual tensor, and it also is approximated by $\beta_3 H_3$. After recursively performing the above operations, they obtain order-K residual quantization:

\begin{equation}
    X=\sum_{i=1}^{K} \beta_i H_i.
\end{equation}

While training the HORQ network, the input tensor can be reshaped to a matrix, and it can be expressed in the form of any order residual quantization. Experiments show that HORQ-Net outperforms XNOR-Net on accuracy on the CIFAR dataset.

ABC-Net \cite{lin2017towards} is another network designed to improve the performance of binary networks. ABC-Net approximates the full-precision weight filter $W$ with a linear combination of $M$ binary filters $B_1, B_2, ... , B_M \in \{+1, -1\}$ such  that $W \approx \alpha_1 \beta_1+ ... +\alpha_M \beta_M$. Those binary filters are fixed as follows:
\begin{equation}
    B_i = F_{u_i}(W) := sign(\bar{W}+u_istd(W)),i=1,2,...,M,
\end{equation}
where $\bar{W}$ and $std(W)$ are the mean and standard derivation of $W$. For activations, ABC-Net employs multiple binary activations to alleviate information loss. Like the binarization weights, the real activation $I$ is estimated using a linear com-bination of $N$ activations $A_1, A_2,..., A_N$ such that $I=\beta_1A_1+ ... +\beta_N A_N$, where
\begin{equation}
\label{eq12}
    A_1, A_2,..., A_N = H_{v_1}(R), H_{v_2}(R),..., H_{v_N}(R).
\end{equation}

$H(R)$ in Eq. \ref{eq12} is a binary function, h is a bounded activation function, $I$ is the indicator function and $v$ is a shift parameter. Unlike the input weights, the parameters $\beta$ and $v$ are both trainable. Without the explicit linear regression, $\beta_n's$ and $v_n's$ are tuned by the network itself during training and fixed for testing. They are expected to learn and utilize the statistical features of full-precision activations.

Ternary-Binary Network (TBN) \cite{WanSLZQSS18} is a convolutional neural network with ternary inputs and binary weights. Based on an accelerated ternary-binary matrix multiplication, TBN uses efficient operations such as XOR, AND, and bit count in standard CNNs and thus provides an optimal tradeoff between memory, efficiency, and performance. Wang et al. \cite{WangH0ZL018} propose a simple yet effective Two-Step Quantization (TSQ) framework by decomposing the network quantization into two steps: code learning and transformation function learning based on the learned codes. TSQ mainly fits 2-bit neural networks.

LBCNN \cite{Juefei-XuBS17} proposes a local binary convolution (LBC), which is motivated by local binary patterns (LBP), a de-scriptor for images rooted in the face recognition community.  The LBC layer has a set of fixed sparse pre-defined binary convolutional filters that are not updated during the training process, as well as a non-linear activation function and a set of learnable linear weights. The linear weights combine the activated filter responses to approximate the corresponding activated filter responses of a standard convolutional layer. The LBC layer affords significant parameter savings commonly 9$x$ to 169$x$ fewer learnable parameters compared to a standard convolutional layer. Furthermore, the sparse and binary nature of the weights also results in up to 169$x$ savings in model size compared to a conventional convolution.

\begin{figure}[!t]
    \centering
    \includegraphics[width=2.5in]{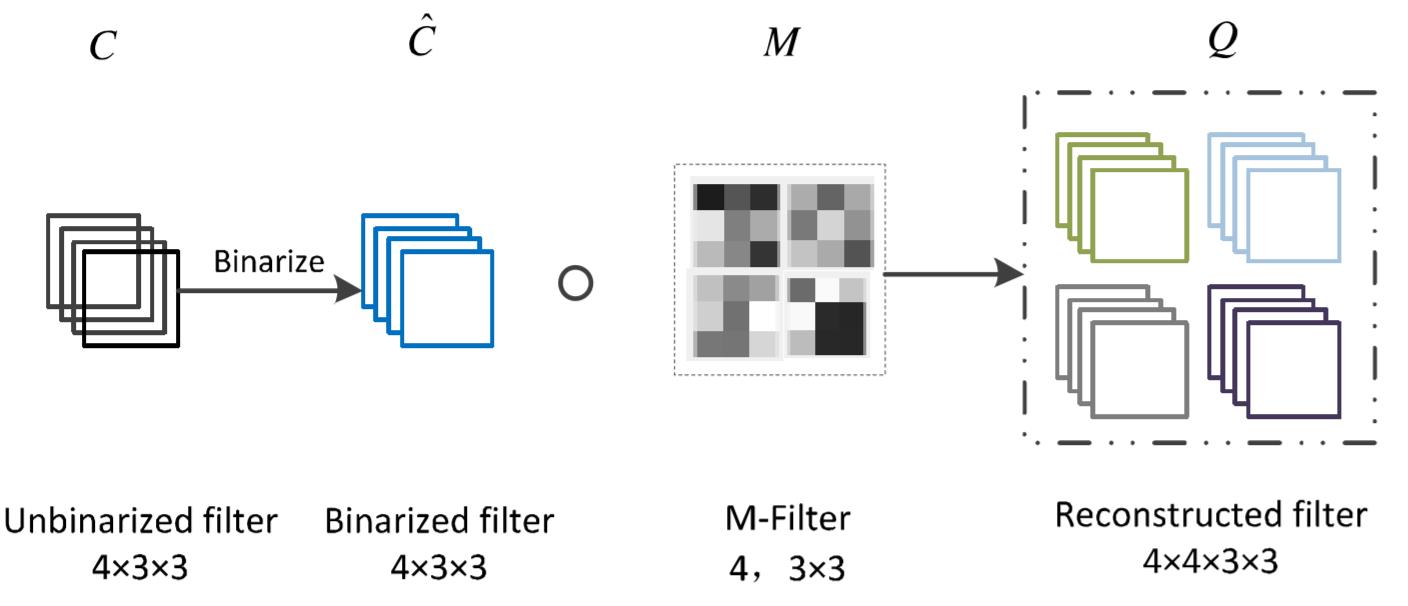}
    \caption{Modulation process based on an M-Filter}
    \label{Figure1}
\end{figure}

MCN \cite{WangZLJH0L18} first introduces modulation filters (M-Filters) to recover the binarized filters. M-Filters are designed to approximate the unbinarized convolutional filters in an end-to-end framework. Each layer shares only one M-Filter and it leads to a significant reduction of the model size. To reconstruct the unbinarized filters, they introduce a modulated process based on the M-Filters and binarized filters. Fig. \ref{Figure1} is an example of the modulation process. In this example, the M-Filter has 4 planes, each of which can be expanded to a 3D matrix according to the channels of the binarized filter. After the $\circ$ operation between the binarized filter and each expanded M-Filter, the reconstructed filter $Q$ is obtained.

\begin{figure}[!t]
    \centering
    \includegraphics[width=2.5in]{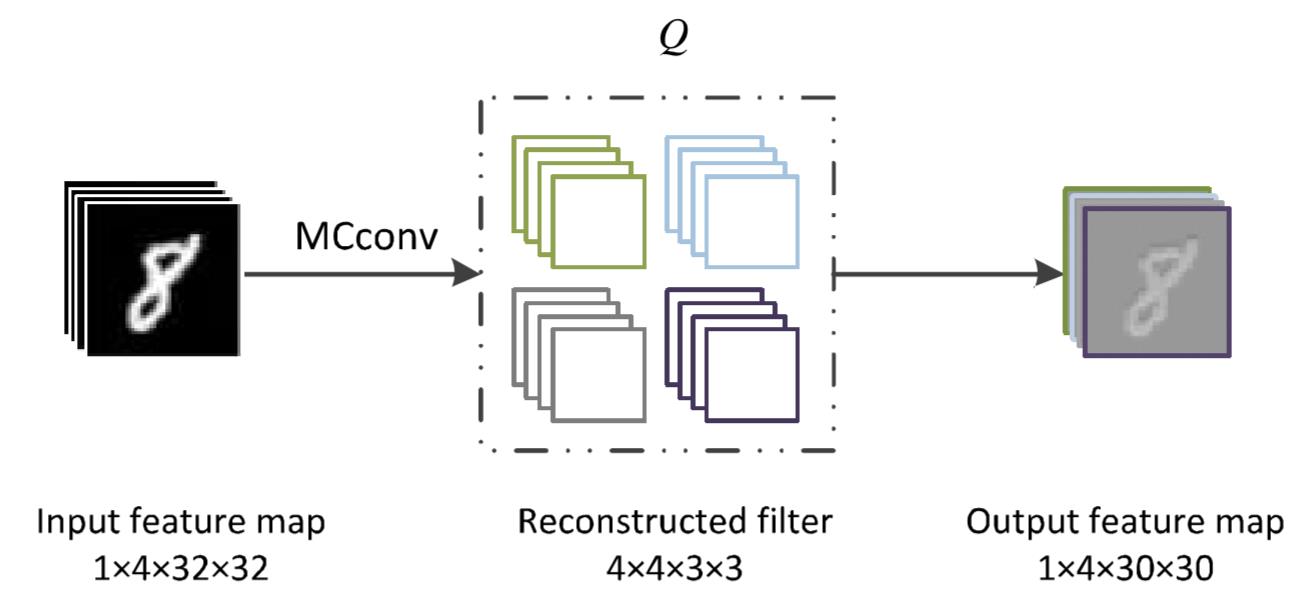}
    \caption{MCNs convolution}
    \label{Figure2}
\end{figure}

As shown in Fig. \ref{Figure2}, the reconstructed filters $Q$ are used to calculate the output feature maps $F$. In Fig. \ref{Figure2}, there are 4 planes, so the number of channels of the feature maps is also 4. By using MCNs convolution, the numbers of input and output channels in every feature maps are the same, allowing the module to be replicated and the MCNs to be easily implemented.

Unlike prior work where the model binarizes each filter independently, Bulat et al. \cite{abs-1904-07852} propose to parametrize the weight tensor of each layer using a matrix or tensor decomposition. The binarization process is performed using the latent parametrization through a quantization function (e.g. sign function) for the reconstructed weights. While the reconstruction is binarized, the computation in the latent factorized space is done in the real domain. This has several advantages.  First, the latent factorization enforces a coupling of filters before binarization, which significantly improves the accuracy of the trained models. Second, during training, the binary weights of each convolutional layer are parametrized using a real-valued matrix or tensor decomposition, while during inference reconstructed (binary) weights are used.

Instead of using the same binary method for weights and activations, Huang et al. \cite{HuangNY19} take the view that the best performance for binarized neural networks can be obtained by applying different quantization methods to weights and activations. They simultaneously binarize the weights and quantizing the activations to reduce bandwidth.

ReActNet \cite{abs-2003-03488} proposes a simple channel-wise reshaping and shifting operation for the activation distribution, which replaces the sign function with ReAct-Sign, and replaces the PReLU function with ReAct-PReLU. Parameters in ReAct-Sign and ReAct-PReLU can be updated.

Compared to XNOR-Net \cite{rastegari2016xnor}, both HORQ-Net \cite{li2017performance} and ABC-Net \cite{lin2017towards} use multiple binary weights and activations. As a result, HORQ-Net and ABC-Net outperform XNOR-Net on binary tasks, but they also increase complexity, which goes against the initial intention of binary neural networks. New neural networks that not only perform better but also retain the advantage of speediness are waiting to be explored. MCN \cite{WangZLJH0L18} and LBCNN \cite{Juefei-XuBS17} proposed new filters while quantizing parameters, and a new loss function is introduced to learn these auxiliary filters.

\subsection{Structural Design}
\label{sec2.4}
The basic structure of networks such as BinaryConnect \cite{courbariaux2015binaryconnect} and BinaryNet \cite{courbariaux2016binarized} is essentially the same as traditional convolutional neural networks, which may not fit the binary process. Some attempts have been made to modify the structure of binary neural networks for better accuracy.

XNOR-Net \cite{rastegari2016xnor} changes the block structure in a typical CNN. A typical block in a CNN contains different layers: 1-Convolutional, 2-BatchNorm, 3-Activation, and 4-Pooling. To further decrease the information loss due to binarization, XNOR-Net normalizes the input before binarization. This ensures the data has zero mean so thresholding at zero minimizes quantization error. The order of the layers in XNOR-Net is shown in Fig. 3.

\begin{figure}[!t]
    \centering
    \includegraphics[width=2.5in]{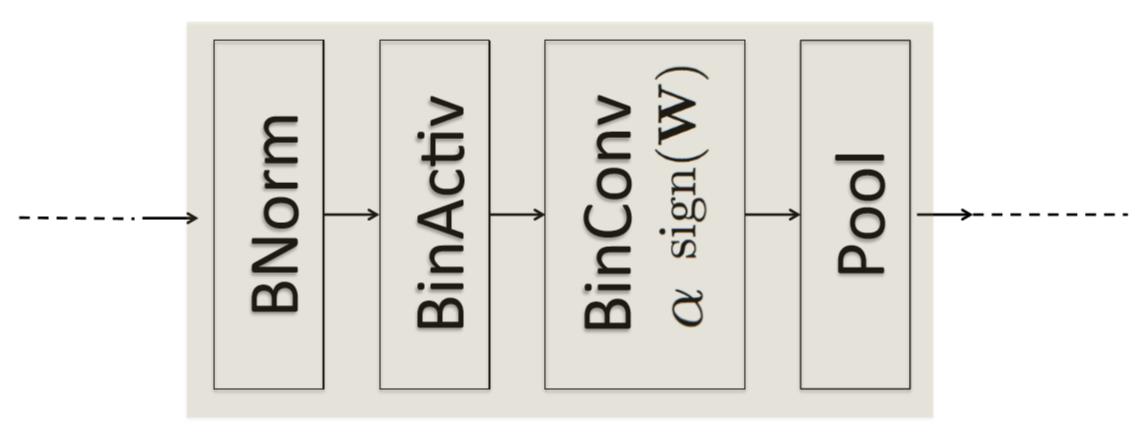}
    \caption{A block in XNOR-Net}
    \label{Figure3}
\end{figure}

\begin{figure}[!t]
    \centering
    \includegraphics[width=1in]{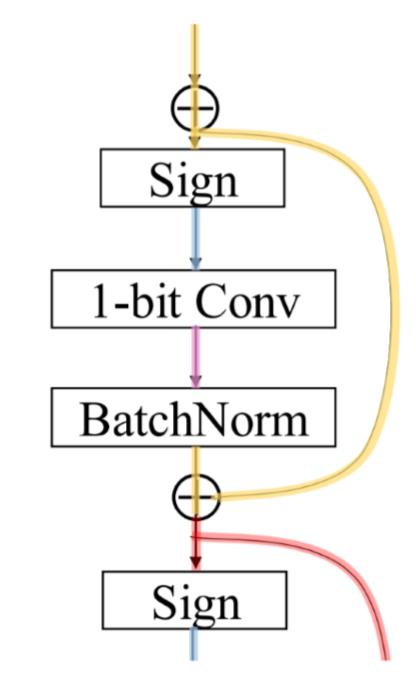}
    \caption{1-bit CNN with shortcut}
    \label{Figure4}
\end{figure}

Bi-real Net \cite{liu2018bi} attributes the poor performance of 1-bit CNNs to its low representation capacity. The representation capacity is defined as the number of all possible configurations of $x$, where $x$ could be a scalar, vector, matrix or tensor. Bi-real Net proposes a simple shortcut to preserve the real activations before the sign function to increase the representation capability of the 1-bit CNN. As shown in Fig. \ref{Figure4}, the block indicates the structure  "Sign $\to$ 1-bit convolution $\to$ batch normalization $\to$ addition operator". The shortcut connects the input activations to the sign function in the current block to the output activations after the batch normalization in the same block, and these two activations are added through an addition operator, and then the combined activations are passed to the sign function in the next block.

The simple identity shortcut significantly enhances the representation capability of each block in the 1-bit CNN. The only additional cost of computation is the addition operation of two real activations without additional memory cost.

\begin{figure}[!t]
    \centering
    \includegraphics[width=1.5in]{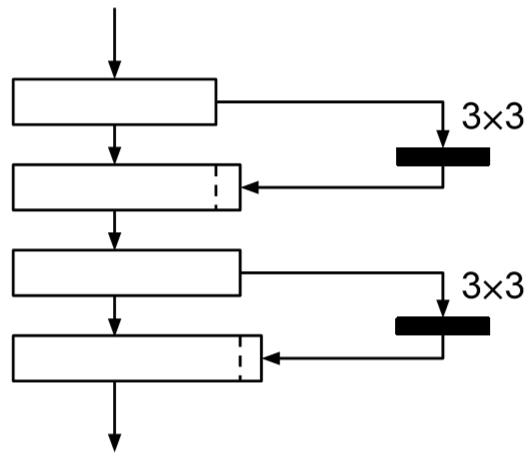}
    \caption{BinaryDenseNet}
    \label{Figure5}
\end{figure}

BinaryDenseNet \cite{abs-1906-08637} designs a new BNN architecture that addresses the main drawbacks of BNNs. DenseNets \cite{HuangLW16a} apply shortcut connections, so new information gained in one layer can be reused throughout the entire depth of the network. This is a significant characteristic that helps to maintain information flow. The bottleneck design in DenseNets reduces the number of filters and values significantly between the layers, resulting in less information flow in the BNNs. These bottlenecks need to be eliminated. Due to the limited representation capacity of binary layers, the DenseNet architecture does not achieve satisfactory performance. This problem is solved by increasing the growth rate or using a larger number of blocks. To keep the number of parameters equal for a given BinaryDenseNet, they halve the growth rate and double the number of blocks at the same time. The architecture of BinaryDenseNet is shown in Fig. \ref{Figure5}

\begin{figure}[!t]
    \centering
    \includegraphics[width=2.5in]{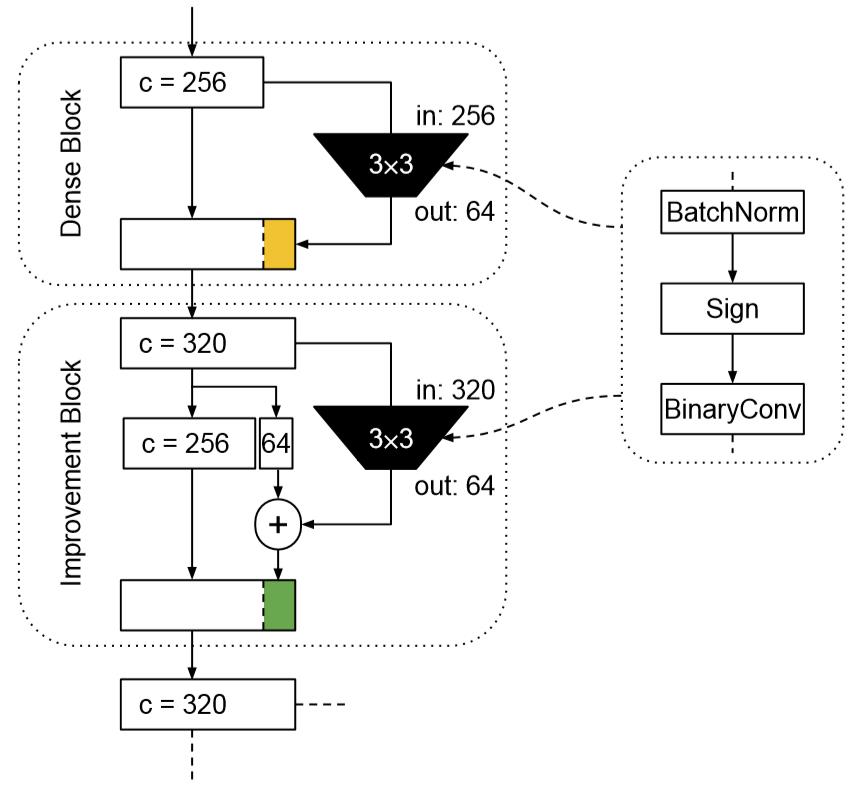}
    \caption{Building blocks of MeliusNet (c denotes the number of channels in the feature map)}
    \label{Figure6}
\end{figure}

MeliusNet \cite{abs-2001-05936} presents a new architecture alternating a DenseBlock, which increases the feature capacity. They also propose an ImprovementBlock, which increases the feature quality. With this method, 1-bit CNNs can match the accuracy of the popular compact network MobileNet-v1 in terms of the model size, the number of operations, and accuracy. The building blocks of MeliusNet are shown as Fig. \ref{Figure6}.

Group-Net \cite{ZhuangSTL019} also improves the performance of 1-bit CNNs through structural design. Inspired by the fact that a floating point number in a computer is represented by a fixed-number of binary digits, Group-Net proposes to de-compose a network into binary structures while preserving its representability, rather than directly doing the quantization via "value decomposition".

Bulat et al. \cite{BulatT17a} is the first to study the effect of neural network binarization on localization tasks, such as human pose estimation and face alignment. They propose a novel hierarchical, parallel, and multi-scale residual architecture that yields a large performance improvement over the standard bottleneck block while maintaining the number of parameters, thus bridging the gap between the original network and its binarized counterpart. The new architecture increases the size of the receptive field as well as the gradient flow.

LightNN \cite{abs-1802-02178} replaces the multiplications with one shift or a constrained number of shifts and adds, which forms a new kind of model. The experiments show LightNN has better accuracy then BNNs with only a slight energy in-crease.

In this section, we listed several works modifying the structure of binary neural networks, which contribute to the better performance or convergence of the network. XNOR-Net and Bi-real Net make some minor adjustments to original networks, while MCN proposes new filters and new convolutional operations. The loss function is also adjusted according to the new filters, which will be introduced in Section \ref{sec2.5}.

\subsection{Loss Design}
\label{sec2.5}
During the optimization of neural networks, the loss function is used to estimate the difference between the real value and the predicted value of a model. Some classical loss functions, such as least square loss and cross-entropy loss are widely used in classification and regression problems. In this section, we will review specific loss function used in binary neural networks.

MCNs \cite{WangZLJH0L18} proposes a novel loss function that considers the filter loss, center loss, and softmax loss in an end-to-end framework. The loss function in MCNs consists of two parts:

\begin{equation}
    L = L_{M} + L_{S}.
\end{equation}

The first part $L_{M}$ is:

\begin{multline}
    L_{M} = \frac{\theta}{2}\sum_{i,l}\big\|C_{i}^{l}-\hat{C_{i}^{l}} \circ M^{l}\big\|^{2} \\ 
    + \frac{\lambda}{2} \sum_{m} \big\|f_m(\hat{C},\vec{M}) - \bar{f}(\hat{C},\vec{M})\big\|^{2},
\end{multline}
where $C$ is the full-precision weights, $\hat{C}$ is the binarized weights, $M$ is the M-Filters defined in Section \ref{sec2.4}, $f_{m}$ denotes the feature map of the last convolutional layer for the $m^{th}$ sample, and $\bar{f}$ denotes the class-specific mean feature map of previous samples. The first entry of $L_{M}$ represents the filter loss, while the second entry calculates the center loss using a conventional loss function, such as softmax loss.

PCNNs \cite{GuLZH0LD19} propose a projection loss for discrete backpropagation. It is the first to define the quantization of the input variable as a projection onto a set to obtain a projection loss. Our BONNs \cite{abs-1908-06314} propose a Bayesian optimized 1-bit CNNs model to significantly improve the performance of 1-bit CNNs. BONNs incorporate the prior distributions of full-precision kernels, features, and filters into a Bayesian framework to construct 1-bit CNNs in a comprehensive end-to-end manner.  They denote the quantization error as $y$ and the full-precision weights as $x$. To minimize the reconstructed error, they maximize $p(x|y)$ to optimize $x$ for quantization. This optimization problem can be converted to a maximum a posteriori since the distribution of $x$ is known. As for feature quantization, the method is the same. Therefore, the Bayesian loss is:

\begin{multline}
    L_{B} = \frac{\lambda}{2} \sum_{l=1}^{l} \sum_{i=1}^{C_{o}^{l}} \sum_{n=1}^{C_{i}^{l}} \{ \big\|\hat{k}_n^{l,i}-w^l \circ k_n^{l,i}\big\|^{2}_2 \\
    + v(k_{n+}^{l,i}-\mu _{i+}^{l})^T (\Psi^{l}_{i+})^{-1} (k_{n+}^{l,i}-\mu _{i+}^{l}) \\
    + v(k_{n-}^{l,i}-\mu _{i-}^{l})^T (\Psi^{l}_{i-})^{-1} (k_{n-}^{l,i}-\mu _{i-}^{l})\\
    vlog(det(\Psi ^{l})) \} + \frac{\theta}{2} \sum_{m=1}^{M} \{ \big\|f_m-c_m\big\|^{2} \\
    + \sum_{n=1}^{N_f} \left[ \sigma _{m,n}^{-2} (f_{m,n}-c_{m,n})^2 + log(\sigma _{m,n}^{2}) \right] \},
\end{multline}
where $k$ is the full-precision kernels, $w$ is the reconstructed matrix, $v$ is the variance of $y$, $\mu$ is the mean of kernels, $\Psi$ is the covariance of kernels, $f_m$ are the features of class $m$, and $c$ is the mean of $f_m$.

Zheng et al. \cite{ZhengDH19} define a new quantization loss between the binary weights and the learned real values where they theoretically prove the necessity of minimizing the weight quantization loss. Ding et al. \cite{abs-1904-02823} propose to use distribution loss to explicitly regularize the activation flow, and develop a framework to systematically formulate the loss. Empirical results show that the proposed distribution loss is robust to the selection of the training hyper-parameters.

Reviewing these methods, they all aim to minimize the error and information loss of quantization, which improves the compactness and capacity of 1-bit CNNs.

\subsection{Neural Architecture Search}
Neural architecture search (NAS) has attracted great attention with remarkable performance in various deep learning tasks. Impressive results have been shown for reinforcement learning (RL) for example\cite{ZophVSL18, ZophL16}. Recent methods like differentiable architecture search (DARTs) \cite{abs-1806-09055} reduce the search time by formulating the task in a differentiable manner. To reduce the redundancy in the network space, partially-connected DARTs (PC-DARTs) was recently introduced to perform a more efficient search without compromising the performance of DARTS \cite{abs-1907-05737}.

In binarized neural architecture search (BNAS) \cite{abs-1911-10862}, neural architecture search is used to search BNNs, and the BNNs obtained by BNAS can outperform conventional models by a large margin. Another natural approach is to use 1-bit CNNs to reduce the computation and memory cost of NAS by taking advantage of the strengths of each in a unified framework \cite{ZhuoZCY0ZD20}. To accomplish this, a Child-Parent (CP) model is introduced to a differentiable NAS to search the binarized architecture (Child) under the supervision of a full precision model (Parent).  In the search stage, the Child-Parent model uses an indicator generated by the child and parent model accuracy to evaluate the performance and abandon operations with less potential. In the training stage, a kernel-level CP loss is introduced to optimize the binarized network. Extensive experiments demonstrate that the proposed CP-NAS achieves a comparable accuracy with traditional NAS on both the CIFAR and ImageNet databases.

Unlike conventional convolutions, BNAS is achieved by transforming all the convolutions in search space $O$ to binarized convolutions. They denote the full-precision and binarized kernels as $X$ and $\hat{X}$ respectively. A convolution operation in $O$ is represented as $B_j  = B_i \otimes \hat{X},$ where $\otimes$ denotes convolution. To build BNAS, one key step is how to binarize the kernels from $X$ to $\hat{X}$, which can be implemented based on state-of-the-art BNNs, such as XNOR or PCNN. To solve this, they introduce channel sampling and operation space reduction into differentiable NAS to significantly reduce the cost of GPU hours, leading to an efficient BNAS.

Shen et al. \cite{ShenHXW19} encode the number of channels in each layer into search space and optimize using the evolutionary algorithm. This approach can identify binary neural architectures for obtaining high precision with as few computation costs as possible.

\subsection{Optimization}
Researchers also explore new training methods to improve the performance of BNNs. These methods are designed to handle the drawbacks of BNNs. Some of them borrow popular techniques from other fields and integrate them into BNNs, while others make changes on the basis of classical BNNs' training, such as improving the optimizer.

Sari et al. \cite{abs-1909-09139} finds that the BatchNorm layer plays a signficant role in avoiding exploding gradients, so the common initialization methods developed for full-precision networks are irrelevant to BNNs. They also break down BatchNorm components to centering and scaling and show only minibatch centering is required. Their work provides valuable information for researches about the training process of BNNs. The experiments of Alizadeh et al. \cite{AlizadehFLG19} show that most of the commonly used tricks in training binary models, such as gradient and weight clipping, are only required during the final stages of the training to achieve the best performance.

XNOR-Net++ \cite{abs-1909-13863} provides a new training algorithm for 1-bit CNNs based on XNOR-Net. Compared with XNOR-Net, this new method fuses the activation and weight scaling factors into a single scaler that is learned dis-criminatively via backpropagation. They also try different ways to construct the shape of the scale factors on the premise that the computational budget remains fixed.

Borrowing an idea from Alternating Direction Method of Multipliers (ADMM), Leng et al. \cite{LengDLZJ18} decouple the continuous parameters from the discrete constraints of the network and cast the original hard problem into several subproblems. These subproblems are solved by extra gradient and iterative quantization algorithms that lead to considerably faster convergence compared to conventional optimization methods.

Deterministic Binary Filters (DBFs) \cite{TsengBFATL18} learn weighted coefficients of predefined orthogonal binary bases instead of the conventional approach which directly learns the convolutional filters. The filters are generated as a linear combination of orthogonal binary codes thus can be generated very efficiently in real-time.

BWNH \cite{HuWC18} trains binary weight networks via hashing. They first reveal the strong connection between inner-product preserving hashing and binary weight networks and show that training binary weight networks can be intrinsically regarded as a hashing problem. They propose an alternating optimization method to learn the hash codes instead of directly learning binary weights.

CI-BCNN \cite{WangLT0019} learns binary neural networks with channel-wise interactions for efficient inference. Unlike existing methods that directly apply XNOR and BITCOUNT operations, this method learns interacted bitcount according to the mined channel-wise interactions. The inconsistent signs in binary feature maps are corrected based on prior knowledge provided by channelwise interactions, so that information of input images is preserved in the forward-propagation of binary neural networks. More specifically, they employ a reinforcement learning model to learn a directed acyclic graph for each convolutional layer, which represent implicit channel-wise interactions. They obtain the interacted bitcount by adjusting the output of the original bitcount in line with the effects exerted by the graph. They train the binary convolutional neural network and the structure of the graph simultaneously.

BinaryRelax \cite{abs-1801-06313} is a two-phase algorithm to train CNNs with quantized weights, including binary weights. They relax the hard constraint into a continuous regularizer via Moreau envelope \cite{moreau1965proximite}, which turns out to be the squared Euclidean distance to the set of quantized weights. They gradually increase the regularization parameter to close the gap between the weights and the quantized state. In the second phase, they introduce the exact quantization scheme with a small learning rate to guarantee fully quantized weights.

\begin{figure}[!t]
    \centering
    \includegraphics[width=3in]{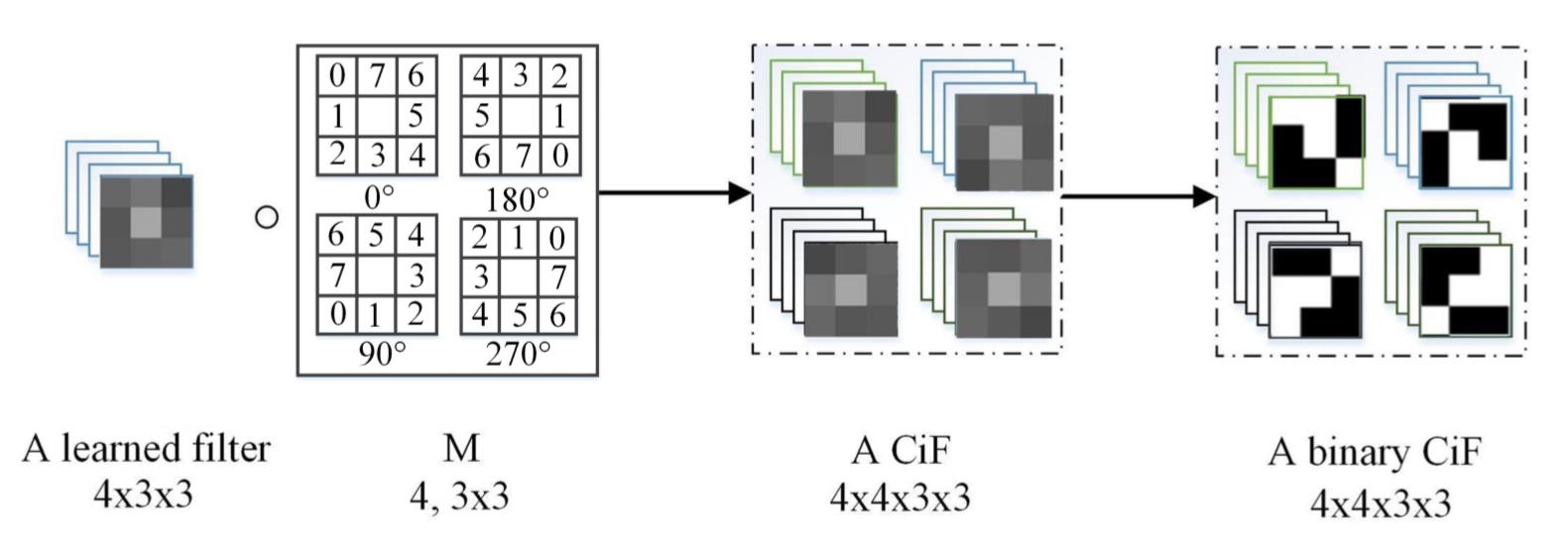}
    \caption{The generation of CiF}
    \label{Figure7}
\end{figure}

CBCNs \cite{abs-1910-10853} propose new circulant filters (CiFs) and a circulant binary convolution (CBConv) to enhance the capacity of binarized convolutional features via circulant backpropagation. A CiF is a 4D tensor of size $K\times K \times H \times H$, generated based on a learned filter, and a circulant transfer matrix $M$. Matrix $M$ here is used to rotate the learned filter at different angles. The original 2D $H \times H$ learned filter is modified to 3D by replicating it three times and concatenating them to obtain 4D CiF, as shown in Fig. \ref{Figure7}. The method can improve the representation capacity of BNNs without changing the model size.

Rectfied binary convolutional networks (RBCNs) \cite{abs-1908-07748} use a GAN to train the 1-bit binary network with the guidance of its corresponding full-precision model, which signficantly improves the performance of 1-bit CNNs. The rectfied convolutional layers are generic and flexible and can be easily incorporated into existing DCNNs such as WideResNets and ResNets.

Martinez et al. \cite{abs-2003-11535} attempt to minimize the discrepancy between the output of the binary and the corresponding real-valued convolution. They propose a real-to-binary attention matching, which is suited for training 1-bit CNNs. They also devise an approach in which the architectural gap between real and binary networks is progressively bridged through a sequence of teacher-student pairs.

Instead of using a pretrained full-precision model, Bethge et al. \cite{abs-1812-01965} directly trains a binary network from scratch, which does not benefit from other common methods. Their implementation is based on the BMXNet framework \cite{YangFBM17}.

Helwegen et al. \cite{abs-1906-02107} take the view that the real-valued latent weights cannot be treated analogously to weights in real-valued networks, while their main role is to provide inertia during training. They introduce the Binary Optimizer (Bop), which is the first optimizer designed for BNNs.

BinaryDuo \cite{abs-2002-06517} proposes a new training scheme for binary activation networks in which two binary activations are coupled into a ternary activation during training. They first decouple a ternary activation to two binary activations. Then the number of weights is doubled after decoupling. To match the parameter size of the decoupled model and the baseline model, they reduce the size of the coupled ternary model. They the update each weight independently so that each of them can find a better value since the two weights no longer need to share the same value.

BENN \cite{ZhuDS19} leverages classical ensemble methods to improve the performance of 1-bit CNNs. While ensemble techniques have been broadly believed to be only marginally helpful for strong classifiers such as deep neural networks, their analysis and experiments show that they are naturally a perfect fit to boost BNNs. The main uses of ensemble strategies are shown in \cite{breiman1996bias,carney1999confidence,oza2001online}.

TentacleNet \cite{abs-1912-10103} is also inspired by the ensemble learning theory. Compared with BENN \cite{ZhuDS19}, TentacleNet takes a step forward, showing that binary ensembles can reach high accuracy with fewer resources.

BayesBiNN \cite{abs-2002-10778} uses a distribution over the binary variable, resulting in a principled approach for discrete optimization. They use a Bernoulli approximation to the posterior and estimate it using the Bayesian learning rule, which is proposed in \cite{khan2019learning}.

\begin{table*}[ht]
    \centering
    \caption{Experimental results of some famous binary methods on ImageNet}
    \noindent\resizebox{\textwidth}{!}{
    \begin{tabular}{cccccccc}
        \toprule
        \multirow{2}{*}{Methods}
        &\multirow{2}{*}{Weights}
        &\multirow{2}{*}{Activations}
        &\multirow{2}{*}{Model}
        &\multicolumn{2}{c}{Binarized Accuracy}  
        &\multicolumn{2}{c}{Full Precision Accuracy} \\
        \cline{5-8} &&&& Top 1 & Top 5 & Top 1 & Top 5 \\
        \hline \hline \\ 
        XNOR-Net \cite{ rastegari2016xnor}&Binary&Binary&ResNet-18  \cite{DBLP:conf/cvpr/HeZRS16} &51.2&73.2&69.3&89.2 \\\hline \\
        ABC-Net \cite{lin2017towards}&Binary&Binary&ResNet-50&70.1&89.7&76.1&92.8\\\hline \\
        LBCNN \cite{Juefei-XuBS17}&Binary&--&--&62.43\footnotemark &--&64.94&--\\\hline \\
        Bi-Real Net \cite{liu2018bi}&Binary&Binary&ResNet-34&62.2&83.9&73.3&91.3\\\hline \\
        PCNN \cite{GuLZH0LD19}&Binary&Binary&ResNet-18  &57.3&80.0&69.3&89.2\\\hline \\
        RBCN \cite{abs-1908-07748}&Binary&Binary&ResNet-18&59.5&81.6&69.3&89.2\\ \hline \\
        BinaryDenseNet \cite{abs-1906-08637}&--&--&--&62.5&83.9&--&--\\ \hline \\
        BNAS \cite{abs-1911-10862}&--&--&--&71.3&90.3&--&--\\ 
        \bottomrule
        \label{table2}
    \end{tabular}}
\end{table*}

\section{Applications}

The success of binary neural networks makes it possible to apply deep learning models to edge computing.  Neural network models have been used in various real tasks with the help of these binary methods including image classification, image classification, speech recognition and object detection and tracking.

\subsection{Image Classification}

Image classification aims to group images into different semantic classes together. Many works regard the completion of image classification as the criterion for the success of binary neural networks. Five datasets are commonly used for image classification tasks: MNIST \cite{netzer2011reading}, SVHN, CIFAR-10 \cite{krizhevsky2009learning}, CIFAR-100, and ImageNet \cite{DBLP:journals/ijcv/RussakovskyDSKS15} . Among them, ImageNet is the most difficult to train and consists of 100 classes of images. Table \ref{table2} shows experimental results of some most popular binary methods on ImageNet.

\footnotetext{13$\times$13 Filter}

\subsection{Speech Recognition}

Speech recognition is a technique or capability that enables a program or system to process human speech. We can make use of binary methods to complete speech recognition tasks in edge computing devices.

Xiang et al. \cite{XiangQ017} applied  binary DNNs to speech recognition tasks. Experiments on both TIMIT phone recognition and a 50-hour Switchboard speech recognition show that binary DNNs can run about 4 times faster than standard DNNs during inference, with roughly 10.0\% relative accuracy reduction.

Zheng et al. \cite{ZhengOSLLWY19} and Yin et al. \cite{YinOZSLLW18} also implement binarized convolutional neural network-based speech recognition tasks.

\subsection{Object Detection and Tracking}

Object detection is the process of finding a target from a scene, while object tracking is follow a target in consecutive frames in a video.

Sun et al. \cite{SunYWXWG18} propose a fast object detection algorithm based on BNNs. Compared to full-precision convolution, this new method results in 62 times faster convolutional operations and 32 times memory saving in theory.

Liu et al. \cite{abs-1908-07748} experiment on object tracking after proposing RBCNs. They use SiamFC Network as the backbone for object tracking and binarize SiamFC as Rectified Binary Convolutional SiamFC Network (RB-SF). They evaluate RBSF on four datasets, GOT-10K \cite{DBLP:journals/corr/abs-1810-11981}, OTB50 \cite{WuLY13}, OTB100 \cite{WuLY15}, and UAV123 \cite{MuellerSG16}, using accuracy occupy (AO) and success rate (SR). The results are shown as Table \ref{table3}.

\begin{table}[ht]
    \centering
    \caption{Results reported in Liu et al. \cite{abs-1908-07748}}
    \noindent
    \begin{tabular}{ccccc}
        \toprule
        Dataset&Index&SiamFC&XNOR&RB-SF \\ \hline
        \multirow{2}{*}{GOT-10K}
        &AO&0.348&0.251&0.327 \\
        \cline{2-5} &SR&0.383&0.230&0.343 \\ \hline \hline
        \multirow{2}{*}{OTB50}
        &Precision&0.761&0.457&0.706 \\
        \cline{2-5} &SR&0.556&0.323&0.496 \\ \hline \hline
        \multirow{2}{*}{OTB100}
        &Precision&0.808&0.541&0.786 \\
        \cline{2-5} &SR&0.602&0.394&0.572 \\ \hline \hline\multirow{2}{*}{UAV123}
        &Precision&0.745&0.547&0.688 \\
        \cline{2-5} &SR&0.528&0.374&0.497 \\
        \bottomrule
        \label{table3}
    \end{tabular}
\end{table}

Yang et al. \cite{YangHF19} propose a new method to optimize YOLO based object tracking deep neural network simultaneously using approximate weight binarization, trainable threshold group binarization activation function, and depth wise separable convolution methods, to greatly reduce computation complexity and model size.

\subsection{Other Applications}

Other applications include face recognition and face alignment.  Face Recognition: Liu et al. \cite{LiuLWCHX20} apply Weight Binarization Cascade Convolution Neural Network to eye localization, which is a field of face recognition. Binary neural networks here help reduce the storage size of the model, as well as speed up calculation.

Face Alignment: Bulat et al. \cite{BulatT17a} test their method on three challenging datasets for large pose face alignment: AFLW \cite{KostingerWRB11}, AFLW-PIFA \cite{JourablooL15}, and AFLW2000-3D \cite{ZhuLLSL16}, reporting in many cases state-of-the-art performance.

\section{Our Work on BNNs}

We have designed several BNNs and 1-bit CNNs. MCN \cite{WangZLJH0L18} was our first work, in which we introduce modulation filters to approximate the unbinarized filters in the end-to-end framework. Based on MCN, we introduced projection convolutional neural networks (PCNNs) \cite{GuLZH0LD19} with a discrete backpropagation via projection. Similar to PCNNs, our CBCNs \cite{abs-1910-10853} aim at improving the backpropagation by improving the representation ability based on a circular backpropagation method. Our RBCNs \cite{abs-1908-07748} and BONNs \cite{abs-1908-06314} on the other hand, improve new model training by changing the loss function and optimization process. RBCNs introduce GAN, while BONNs are based on Bayesian learning.

\begin{figure}[!t]
    \centering
    \includegraphics[width=3in]{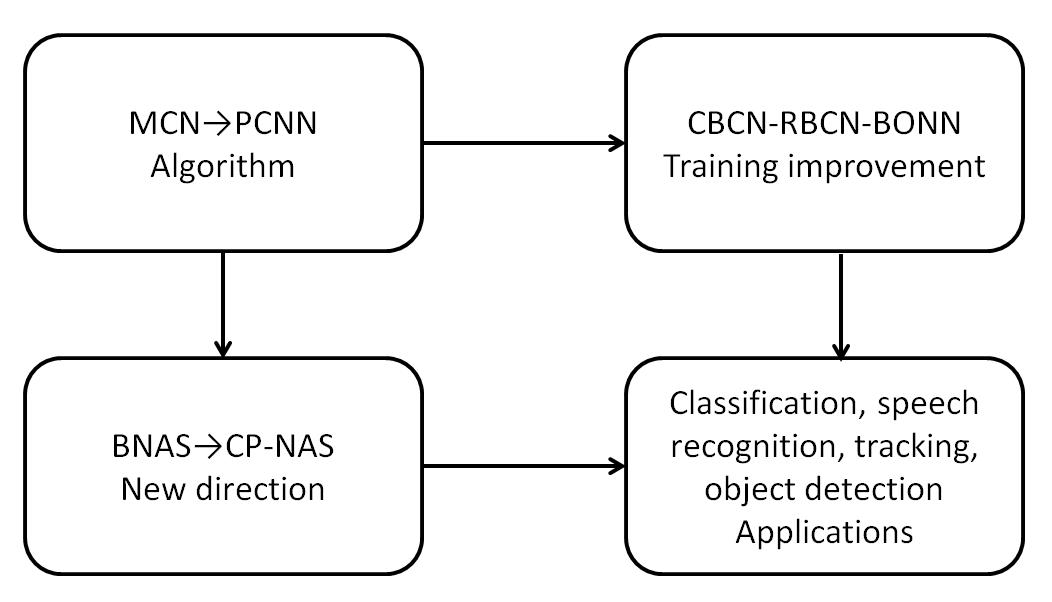}
    \caption{Our Research Agenda on BNNs}
    \label{Figure8}
\end{figure}

Although the performance of BNNs has improved greatly in the past three years, the gap is still large when compared to their full-precision counterparts. One possible solution could come from neural architecture search (NAS), which has led to state-of-the-art performance on many learning tasks. A natural idea is introducing NAS into BNNs, which lead our binarized neural architecture search (BNAS) \cite{abs-1911-10862}. In our BNAS framework, we show that the BNNs obtained by BNAS can outperform conventional models by a large margin. While BNAS only focuses on the kernel binarization, to achieve 1-bit CNNs, our CP-NAS \cite{ZhuoZCY0ZD20} advances this work to binarize both the weights and activations. In CP-NAS, a Child-Parent (CP) model is introduced to a differentiable NAS to search the binarized architecture (Child) under the supervision of a full precision model (Parent). Based on CP-NAS, we achieve a much better performance than conventional binarized neural networks. Our research agenda on BNNs is shown in Fig. \ref{Figure8}

\section{Conclusions and Future Work}
This paper reviewed, compared and contrasted recent advances of binary neural networks (BNNs) for front-end and embedded computing focusing on six aspects including gradient approximation, quantization, structural design, loss design, NAS and optimization. We also introduced several application on speech recognition and computer vision and discuss the future directions for edge computing. Future systems will continue to benefit from performance improvement which are being introduced primarily by NAS related learning algorithms.

\textbf{Acknowledgments.} The work was supported in part by National Natural
Science Foundation of China under Grants 62076016 and 61672079.
This work is supported by Shenzhen Science and Technology Program
KQTD2016112515134654. Baochang Zhang is the corresponding author who is  also with Shenzhen
Academy of Aerospace Technology, Shenzhen, China.

\bibliographystyle{IEEEtran}
\bibliography{final}



% \end{thebibliography}



%\vfill


\end{document}